\pdfoutput=1
\documentclass[11pt]{article}

\usepackage[]{EMNLP2023}

\usepackage{times}
\usepackage{latexsym}
\usepackage{microtype}
\usepackage{epsfig}
\usepackage{amsmath}
\usepackage{amssymb}
\usepackage{booktabs}
\usepackage{algorithmic}
\usepackage{mathtools}
\usepackage{multirow}
\usepackage{multicol}
\usepackage{makecell}

\usepackage{graphicx}

\usepackage[utf8]{inputenc}

\usepackage{multirow}
\usepackage[LAE,LFE,LHE,T1]{fontenc}

\usepackage[farsi,main=english]{babel}

\usepackage{microtype}

\usepackage{inconsolata}

\title{Filling the Gap for Uzbek: Creating Translation Resources for 

Southern Uzbek}

\author{
Mukhammadsaid Mamasaidov\textsuperscript{1} \
Azizullah Aral\textsuperscript{2}\\
\textbf{Abror Shopulatov}\textsuperscript{1,3} \
\textbf{Mironshoh Inomjonov}\textsuperscript{1} \\
        \textsuperscript{1} Tilmoch
        \textsuperscript{2} Academy of Sciences of Afghanistan
        \textsuperscript{3} MBZUAI
        }
        
\begin{document}
\maketitle
\begin{abstract}

Southern Uzbek (uzs) is a Turkic language variety spoken by around 5 million people in Afghanistan and differs significantly from Northern Uzbek (uzn) in phonology, lexicon, and orthography. Despite the large number of speakers, Southern Uzbek is underrepresented in natural language processing. We present new resources for Southern Uzbek machine translation, including a 997-sentence FLORES+ dev set, 39,994 parallel sentences from dictionary, literary, and web sources, and a fine-tuned NLLB-200 model (lutfiy). We also propose a post-processing method for restoring Arabic-script half-space characters, which improves handling of morphological boundaries. All datasets, models, and tools are released publicly to support future work on Southern Uzbek and other low-resource languages.

\end{abstract}

\section{Introduction}

The Southern Uzbek language, spoken by approximately 5 million Uzbeks residing across 14 provinces of Afghanistan, represents a distinct linguistic variety that has developed independently from Northern Uzbek over centuries \cite{ethnologue_uzs}. Uzbek as a whole is classified as a macrolanguage according to ISO 639-3 standards, encompassing multiple related varieties including Northern Uzbek (uzn) spoken primarily in Uzbekistan, and Southern Uzbek (uzs) prevalent in Afghanistan \cite{ethnologue_uzb}. 

This macrolanguage classification recognizes the significant linguistic diversity within the broader Uzbek language family, where individual varieties have developed distinct phonological, lexical, and grammatical features due to geographical separation and contact with other languages. As part of the global Uzbek population exceeding 34 million people, Southern Uzbek is recognized in Afghanistan’s Constitution as a potential third official language in regions where it is the majority language, in addition to Pashto and Dari. \cite{afghanistan_constitution_2004}

Southern Uzbek functions as a fully developed literary language that meets the demands of literature, art, culture, and science. It maintains active presence across multiple domains including technology, education, diplomacy, banking, and commerce. The language is taught in Southern Uzbek departments at seven national universities in Afghanistan and serves as the medium of instruction in 970 schools distributed across provinces: 9 schools in Badakhshan, 80 in Balkh, 450 in Faryab, 50 in Samangan, 300 in Sar-e-Pol, and 80 in Takhar. \cite{unforgettable2020}

International media outlets including BBC, Radio Free Europe/Radio Liberty (Ozodlik), Voice of America, Voice of Iran, TRT Avaz, and Sputnik actively broadcast in Southern Uzbek, alongside Afghan media channels such as Oyna, Botur, Almas, Orzu, Nur, Oriano, Kalid, and National Radio and Television. The language maintains expanding digital presence across major online platforms including Wikipedia, Google, Facebook, and other social networks.

Despite this linguistic vitality, Southern Uzbek remains underrepresented in natural language processing technologies. Major translation platforms like Google Translate \cite{GoogleTranslate} currently provide limited or no support for this language variety, highlighting the critical need for dedicated computational resources. As a low-resource language with unique characteristics distinct from Northern Uzbek, Southern Uzbek presents significant challenges for machine translation systems.

This study, conducted as part of the Open Language Data Initiative (OLDI) shared task, addresses these challenges by developing specialized neural machine translation models for Southern Uzbek. Our contributions parallel recent advances in low-resource language processing and include:

\begin{enumerate}
    \item A FLORES+ dev dataset translated to Southern Uzbek containing 997 sentences
    \item Parallel corpora for various language pairs with Southern Uzbek
    \item Open-sourced fine-tuned neural models for Southern Uzbek translation
    \item Comprehensive evaluation against existing baselines
\end{enumerate}

Our research aims to advance machine translation capabilities for Southern Uzbek, contributing to the larger OLDI objective of expanding linguistic diversity in NLP technologies for underrepresented language varieties.

% \begin{table*}[t]
% \centering
% \small
% \begin{tabular}{|p{0.48\textwidth}|p{0.48\textwidth}|}
% \hline
% \textbf{English} & \textbf{Karakalpak} \\
% \hline
% According to Japan's nuclear agency, radioactive caesium and iodine has been identified at the plant. & Yaponiya yadro agentligi maǵlıwmatlarına kóre, stanciyada radioaktiv ceziy hám yod bar ekenligi anıqlanǵan. \\
% \hline
% The result of plotting analysis will be posted to a public website. & Syujet analiziniń nátiyjesi ǵalabalıq veb-saytqa jaylastırıladı. \\
% \hline
% The station's web site describes the show as "old school radio theater with a new and outrageous geeky spin!" & Stanciya veb-saytında show "jańa hám ádettegiden basqasha ájáyıp aylandıratuǵın eski mektep radio teatrı!" dep táriyiplenedi. \\
% \hline
% \end{tabular}
% \caption{Examples from the FLORES+ dataset for English-Karakalpak language pair}
% \label{tab:flores-examples}
% \end{table*}

\section{Linguistic Background}

\subsection{Historical Development}

Southern Uzbek belongs to the Turkic language family, specifically derived from the Karluk-Chigil-Uyghur dialectal group with partial influences from the Kipchak and Oghuz branches. The language represents the contemporary form of a literary tradition spanning over a millennium, with historical continuity traceable through classical poets including Khwarizmi, Lutfi, Atayi, Sakkaki, Navoi, Babur, lutfiy, and Ogahi. Notably, while these historical figures did not identify themselves as ``Uzbek'', they wrote in a language that forms the foundation of modern Southern Uzbek, demonstrating the language's independent development into a mature linguistic system. \cite{habibi2021brief}

% Contemporary Southern Uzbek speakers can comprehend classical literary works without glossaries, illustrating the language's historical continuity and stability. While Southern and Northern Uzbek share fundamental grammatical structures, they exhibit notable lexical differences that have emerged through centuries of independent development.

Historically, Southern Uzbek served as the administrative and literary language for major dynasties including the Yaftids, Kushans, Ghaznavids, Seljuks, Timurids, and Mughals, who governed territories across Afghanistan and India for centuries using this language and established profound cultural legacies. \cite{tursunov1982}

\subsection{Writing System}

Southern Uzbek employs the Arabic script, which has served as the official writing system for Afghan languages for over a thousand years. This orthographic system presents unique challenges and characteristics that distinguish it from Latin-based Northern Uzbek.

\begin{figure}[htbp]
    \centering
    \includegraphics[width=\columnwidth]{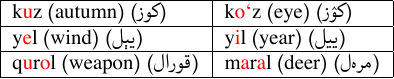}
    \caption{Vowel differences in Southern Uzbek}
    \label{fig:vowel-ambiguity}
\end{figure}

The Arabic-based script includes only three vowel letters: \textRL{ا} (a/o), \textRL{و} (u/o‘), and \textRL{ی} (i/y). This limited vowel representation often misleads learners into believing that Uzbek contains only three vowel sounds. However, vowel quality distinctions become evident in minimal pairs such as shown in Figure \ref{fig:vowel-ambiguity}.

% \begin{itemize} 
% % (\textRL{کۉز})
% % (\textRL{ییل})
%     \item ko‘z  vs. kuz (\textRL{کوز})
%     \item yil  vs. yel (\textRL{یېل})  
%     \item maral (\textRL{مره‌ل}) vs. qurol (\textRL{قورال})
% \end{itemize}

Standard Uzbek contains six primary vowels (with additional dialectal variants), yet the Arabic script lacks direct representation for half of them. These vowels require indication through diacritical marks (fatha, damma, kasra), which are frequently omitted in practical writing, thereby complicating accurate reading and pronunciation.

\begin{figure}[htbp]
    \centering
    \includegraphics[width=\columnwidth]{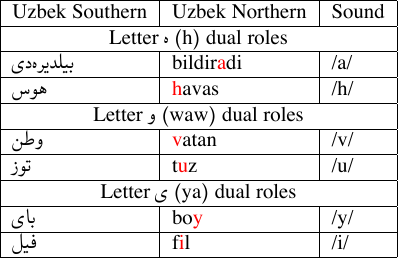}
    \caption{Dual letters in Southern Uzbek}
    \label{fig:dual-letters}
\end{figure}

Additional complexity arises from the dual functionality of certain letters. The Arabic letter \textRL{ه} (h) functions both as vowel and consonant. Similarly, letters \textRL{و} and \textRL{ی} (waw and ya) serve dual roles as vowels and consonants (``v'' and ``y'') depending on context as illustrated in Figure \ref{fig:dual-letters}.

\begin{figure}[htbp]
    \centering
    \includegraphics[width=\columnwidth]{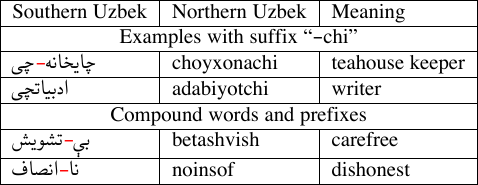}
    \caption{Examples of standardized Southern Uzbek Arabic-script orthography showing mandatory half-space (zero-width non-joiner, U+200C) placement. Red marks indicate the location of half-spaces in suffixation after vowel-final stems and in prefix attachment.}
    \label{fig:zwnj-examples}
\end{figure}

Arabic and Persian loanwords maintain their original orthographic forms, typically without vowel markings.

% :
% \begin{itemize}
%     \item sobun - \textRL{صابون}
%     \item solon - \textRL{صالون}
%     \item to‘fon - \textRL{طوفان}
%     \item to‘ra - \textRL{طوره}
% \end{itemize}

\subsection{Morphological Structure}

Southern Uzbek exhibits rich agglutinative morphology characteristic of Turkic languages. The language employs extensive suffixation systems that can be classified into various functional categories:

\begin{itemize}
    \item Nominalizers (noun-forming suffixes)
    \item Adjectival suffixes  
    \item Verb formers
    \item Tense and aspect markers
    \item Other functional and derivational affixes
\end{itemize}

Standardized orthographic rules govern affix attachment in Southern Uzbek Arabic script. A fundamental principle distinguishes between suffixes attached to vowel-final versus consonant-final stems ( -chi, -chilik, -lik, -li, etc.).

\begin{figure*}[t]
    \centering
    \includegraphics[width=\textwidth]{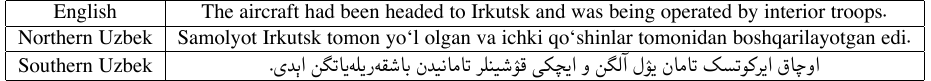}
    \caption{Example from the FLORES+ dataset in English, Northern Uzbek and Southern Uzbek.}
    \label{fig:flores-example}
\end{figure*}

These suffixes require half-space (also known as zero-width non-joiner, U+200C, also found in Farsi) separation when attached to stems ending in vowels (represented by Arabic letters \textRL{ه}، \textRL{و}، \textRL{ا}), while connecting directly to consonant-final stems. 

Southern Uzbek also employs prefixes, commonly found in Persian or Arabic loanwords, for forming adjectives or adverbs. These prefixes (be-, no-, xo‘sh-, ser-, ba-, ham-, bad-) are written with half-space separation, as shown in Figure \ref{fig:zwnj-examples}.

\subsection{Contemporary Status and Challenges}

Despite its historical significance, Southern Uzbek has faced political marginalization over the past three centuries, with Turkic peoples in Afghanistan being sidelined in governance and education. Progress began in the 1970s when Uzbek parliamentary representatives secured broadcasting rights on Afghan national radio. The 1978 rise of the People's Democratic Party marked further advancement with the publication of the Yulduz newspaper in Southern Uzbek, establishment of Uzbek Language and Literature departments, and expansion of Uzbek-medium education. \cite{aral2025navoi}

The 2001 democratic reforms in Afghanistan formally granted Southern Uzbek official status, recognizing its role in Afghan multilingual society. However, challenges remain in standardizing orthographic practices and developing computational resources for this linguistically rich but technologically underrepresented variety.

\section{Related Work}
Machine translation for low-resource languages has gained significant attention, with researchers exploring various approaches from data augmentation to multilingual transfer learning. \citet{dale-2022-first} developed the first neural MT system for Erzya, a low-resource Uralic language, demonstrating how extensive data mining from diverse sources (Bible texts, dictionaries, digitized books) can yield functional translation systems despite limited parallel data. Similarly, \citet{p-m-etal-2024-mtnlp} focused on low-resource Indic languages by fine-tuning multilingual models and employing back-translation with careful quality filtering, showing that selective data augmentation can improve performance when synthetic data is judiciously filtered.

\citet{goyle2023neuralmachinetranslationlow} systematically evaluated strategies for compensating data scarcity in languages like Sinhala, Nepali, Khmer, and Pashto. They found that combining back-translation with focal loss yields substantial improvements, particularly when leveraging large monolingual corpora and transfer learning from related high-resource languages.

Recent advances in large language models have also shown promise for low-resource translation tasks. Commercial LLMs like GPT-4 and Claude demonstrate multilingual capabilities that extend to languages not explicitly included in their training data, offering competitive performance through few-shot learning approaches.

Despite these advances, Southern Uzbek remains largely unexplored in computational linguistics. While Northern Uzbek has received some attention in multilingual models like NLLB \cite{nllb-22} and MADLAD-400 \cite{kudugunta2024madlad}, the Southern Uzbek has been left behind. Our work represents the first dedicated effort to develop neural translation resources for this variety of Uzbek.

\section{Datasets}
\subsection{FLORES+ Dev Dataset}
This study introduces the Southern Uzbek FLORES+ dev dataset, comprising 997 sentences translated from English to Southern Uzbek (see Figure \ref{fig:flores-example}).

The dataset was developed under the Open Language Data Initiative (OLDI) framework. One native Southern Uzbek linguist was responsible for the translation process, with subsequent post-review process to ensure linguistic accuracy and cultural appropriateness. All Southern Uzbek translations strictly adhere to the Arabic script orthographic conventions, including proper implementation of half-space characters (U+200C) for morphological boundaries as described in Section 2.3.

Given the complexity of Arabic script representation and the morphologically rich nature of Southern Uzbek, particular attention was paid to maintaining consistent orthographic standards throughout the translation process. The translation process followed standardized conventions for affix attachment, vowel representation, and proper handling of Arabic and Persian loanwords within the Southern Uzbek linguistic system.
\subsection{Training Data}
The training dataset comprises diverse parallel corpora sourced from three primary domains, totaling 39,994 sentence pairs across multiple language combinations:
\begin{enumerate}
\item \textbf{Dictionary Entries (1,550 pairs)}: Parallel dictionary entries mapping Northern Uzbek to Southern Uzbek lexical items \cite{aral2024phrasebook}. These entries provide direct lexical correspondences and serve as high-quality alignment data for closely related language varieties.
\item \textbf{Literary Corpus (35,865 pairs)}: Parallel sentences extracted through careful alignment from 27 selected books available in both Northern and Southern Uzbek variants. This corpus represents the largest component of our training data and captures literary register variations, complex syntactic structures, and cultural terminology.
\item \textbf{Web-sourced Content (2,579 pairs)}: Parallel sentences of English-Southern Uzbek mined from official government websites and reliable online resources. This component provides contemporary usage patterns and domain-specific terminology from governmental and institutional contexts.
\end{enumerate}

\begin{table*}[t]
\centering
% \small
\begin{tabular}{l|cccc}
\hline
\textbf{Model} & \textbf{uzs-en} & \textbf{uzs-uzn} & \textbf{eng-uzs} & \textbf{uzn-uzs} \\
\hline
gpt-4.1 & 24.90 / 53.42 & 2.634 / 3.657 & 0.48 / 9.49 & 1.42 / 21.55 \\
gemini-2.0-flash-001 & \textbf{32.81 / 58.80} & \textbf{62.45} / 73.67 & 1.59 / 24.47 & 6.96 / 41.11 \\
claude-sonnet-4 & 22.25 / 51.46 & 59.18 / \textbf{83.63} & 0.68 / 15.38 & 2.62 / 28.85 \\
nllb-200-600M & 3.73 / 23.88 & 4.14 / 27.02 & - & - \\
Google Translate & 9.56 / 33.58 & 5.13 / 33.19 & - & - \\
madlad400-3b-mt & 2.95 / 23.26 & 0.19 / 1.41 & - & - \\
\hline
lutfiy (no half-space fix) & \multirow{2}{*}{11.26 / 34.39} & \multirow{2}{*}{53.48 / 78.54} & 1.33 / 25.43 & 25.99 / 66.44 \\
lutfiy (with half-space fix) & & & \textbf{1.58 / 26.61} & \textbf{34.31 / 71.11} \\
\hline
\end{tabular}
\caption{Evaluation of several models on sacreBLEU/chrF++ across various language pairs involving English, Northern Uzbek (uzn) and Southern Uzbek (uzs).}
\label{tab:uzbek-translation-performance}
\end{table*}
% The compilation strategy prioritized domain diversity while maintaining translation quality. Dictionary entries ensure lexical coverage, literary texts provide syntactic complexity and cultural authenticity, while web-sourced materials contribute contemporary usage patterns and specialized terminology.
\subsection{Data Mining Process}
The sentence alignment process presented unique challenges due to Southern Uzbek's underrepresentation in existing multilingual models. Our alignment methodology employed a two-stage approach to maximize extraction efficiency.

For literary corpus alignment, we initially applied LaBSE embeddings \cite{feng2020language} directly to the original Arabic script texts. While LaBSE does not include Southern Uzbek in its training data, the model demonstrated limited alignment capability, likely due to shared vocabulary with other Turkic languages in the embedding space.

To improve alignment quality, we implemented a transliteration-based enhancement strategy. Southern Uzbek texts were transliterated from Arabic to Latin script using rule-based conversion scripts\footnote{\url{https://github.com/tahrirchi/uzs-scripts}}, which enabled more effective cross-lingual embedding alignment. This transliteration approach yielded a 40\% more successfully aligned sentence pairs compared to direct Arabic script processing.

The sentence alignment methodology follows established practices from low-resource language processing \cite{dale-2022-first}. We utilize LaBSE to generate embeddings for each potential sentence pair, calculate cosine similarity between embeddings, and adjust similarity scores using length ratios.

For web-sourced English-Southern Uzbek data, we employed a reverse translation verification approach. Southern Uzbek sentences were translated to English using Gemini-2.0-Flash, followed by LaBSE-based alignment between original English content and back-translated English. This process underwent manual review to ensure translation quality and semantic fidelity.

A notable preprocessing challenge emerged regarding half-space character consistency. Due to OCR limitations and editorial inconsistencies in source materials, half-space characters (U+200C) were frequently omitted, incorrectly rendered as full spaces, or merged with adjacent characters. While this issue complicates training data quality, we address it through post-processing correction mechanisms described in Section 4.

\section{Translation Experiments}
\subsection{Model Training}
Our experimental framework employed the nllb-200-distilled-600M model as the foundation for Southern Uzbek machine translation development. e maintained the original tokenizer configuration, leveraging the model's existing multilingual capabilities for Turkic language processing.
\subsubsection{Training Configuration}
For the training process we employed the Adafactor \cite{shazeer2018adafactor} optimizer paired with a learning rate of \(1\times 10^{-4}\) following a constant schedule and 1000 warmup steps. A weight decay of \(1\times 10^{-3}\) was applied, and the batch size was set to 32 due to GPU memory constraints. The maximum sequence length was limited to 128 tokens, and training was conducted for 5000 steps, corresponding to approximately 2–3 epochs. All experiments were run on a single A100 40GB GPU. The Adafactor optimizer was chosen for its memory efficiency and proven effectiveness in transformer fine-tuning scenarios, while the conservative learning rate and weight decay values were selected to mitigate overfitting given the small size of the training dataset.

% \subsubsection{Training Configuration}
% The training process utilized the following hyperparameter configuration optimized for low-resource language fine-tuning:
% \begin{itemize}
% \item \textbf{Optimizer}: Adafactor with scale\_parameter=False, relative\_step=False
% \item \textbf{Learning Rate}: 1e-4 with constant schedule and 1000 warmup steps
% \item \textbf{Weight Decay}: 1e-3
% \item \textbf{Batch Size}: 32 (constrained by GPU memory limitations)
% \item \textbf{Maximum Sequence Length}: 128 tokens
% \item \textbf{Training Steps}: 5000 (approximately 2-3 epochs)
% \item \textbf{Hardware}: Single A100 40GB GPU
% \end{itemize}
% The Adafactor optimizer was selected for its memory efficiency and demonstrated effectiveness in transformer fine-tuning scenarios. The relatively conservative learning rate and weight decay values were chosen to prevent overfitting given the limited training data size.

\subsubsection{Model Variant}

We fine-tuned nllb-200-distilled-600M \cite{nllb-22} model on the complete 39,994 sentence pair corpus. Our model called \textbf{lutfiy}\footnote{Lutfi, a 15th-century Central Asian poet
} maintains the original NLLB tokenizer and vocabulary, relying on existing Turkic language representations for Southern Uzbek processing.
\subsubsection{Half-Space Post-Processing}
A critical technical challenge emerged regarding the handling of half-space characters. The NLLB SentencePiece \cite{kudo2018sentencepiece} tokenizer normalizes half-space characters (U+200C) to regular spaces during preprocessing, preventing the model from learning proper morphological boundary representation. This problem affects not only Southern Uzbek but also extends to other languages requiring half-space characters, including Persian \cite{doostmohammadi-etal-2020-joint}.

To address this limitation, we developed a character-level n-gram post-processing model that predicts half-space insertion positions. The model was trained on a small set of  training data with corrected half-space characters. It analyzes character sequences and applies statistical rules to determine whether half-spaces should follow specific vowel endings in morphologically complex constructions.

% The n-gram model operates by:
% \begin{enumerate}
% \item Analyzing character-level patterns surrounding potential half-space positions
% \item Computing frequency statistics for half-space occurrence after vowel-final stems
% \item Applying probabilistic rules based on morphological suffix patterns
% \end{enumerate}
This approach provides a practical solution to the tokenizer normalization problem while maintaining compatibility with existing NLLB infrastructure. The post-processing correction mechanism is made publicly available alongside our trained models\footnote{\url{https://huggingface.co/tahrirchi/lutfiy}}.

\subsection{Evaluation Framework}
Model performance was assessed using two widely adopted metrics for translation tasks: \textbf{sacreBLEU} \cite{post-2018-call}, a standardized BLEU implementation that ensures consistent n-gram precision measurement across experiments, and \textbf{chrF++} \cite{popovic2017chrf++}, a character-level F-score metric that is particularly well-suited for evaluating morphologically rich languages such as Southern Uzbek. All results are reported on the FLORES+ dev set, enabling comparability with other low-resource language initiatives under the OLDI framework.

\section{Results and Discussion}

Our evaluation on the FLORES+ Southern Uzbek dev set reveals several key insights into the performance of various translation approaches. The results, presented in Table \ref{tab:uzbek-translation-performance}, demonstrate significant performance variations across different model architectures and translation directions.

Notably, large language models exhibit superior performance in understanding Southern Uzbek content, particularly in \textbf{uzs-*} directions. Gemini-2.0-Flash achieves the highest scores for uzs-en translation (32.81 BLEU/58.80 chrF++), while Claude-Sonnet-4 excels in uzs-uzn translation quality (83.63 chrF++). This suggests that LLMs' extensive multilingual pretraining enables effective comprehension of low-resource language varieties, even without explicit training on Southern Uzbek data. In contrast, traditional MT systems like Google Translate and specialized multilingual models (NLLB-200-600M, MaLLaD400) demonstrate substantially lower performance, highlighting the challenges these architectures face with underrepresented languages.

However, our fine-tuned \textbf{lutfiy} model demonstrates clear advantages in generation tasks. For translation \textbf{into} Southern Uzbek (en-uzs and uzn-uzs), our model consistently outperforms all baselines, achieving 1.58 BLEU/26.61 chrF++ for en-uzs and 34.31 BLEU/71.11 chrF++ for uzn-uzs directions. This validates our approach of fine-tuning on domain-specific parallel corpora, as the model learns proper Southern Uzbek generation patterns that generic LLMs cannot replicate effectively.

The impact of our half-space post-processing correction is particularly evident in the uzn-uzs translation pair. While chrF++ scores show modest improvements (from 66.44 to 71.11), BLEU scores increase dramatically (from 25.99 to 34.31), representing a 32\% relative improvement. This substantial BLEU gain with stable chrF++ performance indicates that the half-space correction primarily addresses tokenization boundary issues rather than fundamental translation errors. Since BLEU relies on exact n-gram matches, incorrect half-space placement can artificially deflate scores even when the underlying translation quality remains high.

For the closely related uzs-uzn translation direction, Gemini-2.0-Flash demonstrates exceptional generation capability (62.45 BLEU), significantly outperforming our specialized model (53.48 BLEU). This suggests that LLMs may be particularly effective at cross-dialectal translation within the same language family, possibly due to their ability to capture subtle linguistic variations during pretraining.

These findings highlight complementary strengths between LLMs and specialized fine-tuned models: while LLMs excel at understanding and translating from Southern Uzbek, targeted fine-tuning proves essential for high-quality generation into Southern Uzbek, particularly for morphologically complex constructions requiring proper orthographic conventions.

\section{Conclusion}
Our study presents the first comprehensive neural machine translation resources for Southern Uzbek, addressing a significant gap in computational linguistics for this underrepresented Turkic variety. Our key contributions include:
\begin{enumerate}
\item Creation of a 997-sentence FLORES+ dev dataset for Southern Uzbek
\item Development of 39,994 parallel sentence pairs across multiple language combinations (uzs-uzn, uzs-en)
\item Fine-tuned NLLB-200 model (lutfiy) optimized for Southern Uzbek translation
\item Post-processing methodology for Arabic script half-space character restoration
\item Open-sourced datasets, models, and evaluation tools
\end{enumerate}
Future work will focus on expanding dataset coverage through additional literary sources and government documents, exploring data augmentation techniques using large language models, and developing more sophisticated orthographic normalization approaches for Arabic script processing.
\section{Limitations}
Several limitations constrain our current approach. The training dataset size of \textasciitilde40K sentence pairs, while substantial for a low-resource language, may limit generalization across diverse domains. Our heavy reliance on literary sources potentially biases the model toward formal registers, possibly affecting performance on conversational or technical content. The half-space post-processing solution, while effective, represents a workaround rather than addressing the underlying tokenizer limitations. Additionally, our evaluation relies primarily on automatic metrics, which may not fully capture translation quality nuances for morphologically complex languages like Southern Uzbek. Human evaluation studies would provide more comprehensive quality assessment.

\section{Acknowledgements}
We thank the Open Language Data Initiative (OLDI) for supporting this research and David Dale for his valuable guidance throughout the project. The authors thank the Google for Startups Program for providing the computational resources that made this research possible.

\clearpage

% Entries for the entire Anthology, followed by custom entries
\bibliography{anthology,custom}
\bibliographystyle{acl_natbib}

\clearpage

% \onecolumn

% \appendix

% \section{Prompt for translating from Russian to English using Claude-3.5-sonnet}
% \label{sec:claude-prompt}

% \begin{verbatim}

% You are a professional translator specializing in Russian to English translations. 
% Your task is to translate the given Russian text into English with the highest level 
% of accuracy, preserving the original meaning and context. Use proper grammar, 
% punctuation, and idiomatic expressions appropriate for English speakers. 
% Do not include any additional explanations or commentary; provide only the translated text.

% Russian: {sent}
% English:

% \end{verbatim}

\end{document}